\documentstyle[12pt]{article}
\title{
On Irrelevance of Attributes in  Flexible Prediction%\\
}

\newcommand{\gh}{\tt}
\newcommand{\Bem}[1]{}

\author{Mieczyslaw A. Klopotek, Andrzej Matuszewski \\  
         Institute of Computer Science, 
         Polish Academy of Sciences\\
         PL 01-237 Warsaw, 21 Ordona St.,\\
         Fax: (48-22)  37-65-64, Phone: (48-22) 36-28-85,\\
         e-mail: klopotek , amat{@}wars.ipipan.waw.pl}

\date{}
\begin{document}

\maketitle
\pagestyle{empty}
\thispagestyle{empty}
\begin{abstract}
This paper analyses properties of conceptual hierarchy obtained via 
incremental concept formation method called "flexible prediction" in order to
determine what kind of "relevance" of participating attributes may be
requested for  meaningful conceptual hierarchy. The impact of
selection of simple and combined attributes, of scaling and of distribution
of individual attributes
and of correlation strengths among them  is investigated. \\
Paradoxically, both: attributes weakly and strongly related with other
attributes have deteriorating impact onto the overall classification. 
Proper construction of derived  attributes as well as  selection of scaling of
individual attributes strongly influences the obtained concept hierarchy.
Attribute density of distribution seems to influence the classification 
weakly\\
 It seems
also, that concept hierarchies (taxonomies) reflect  a compromise between
the data and our interests in   some objective truth about the data.\\
To obtain classifications more suitable for one's purposes, breaking the
symmetry among attributes (by dividing them into dependent and independent
and applying differing evaluation formulas for their contribution) is
suggested.\\
Both continuous and discrete variables are considered. Some methodologies for
the former are considered.\\

\end{abstract}

%\newpage

\section{Introduction}

Most knowledge discovery methods strive to obtain a kind of explicit (by
dividing the set of observed objects into sets) or implicit (by deriving a
dependence function between attributes of objects) "clustering" of objects
based on a kind of "similarity" of them. There exist numerous statistical
clustering methods (single linkage, k-means, \dots) as well as ones
elaborated within artificial intelligence community (conceptual clustering,
incremental concept formation, \dots and many others). Many seem to share one
common feature: complaints about impact of "irrelevant"  attributes which
worsen the satisfaction with the resulting clustering.%\\

This paper tries to shed some light into the question what that "irrelevance"
may mean. To make the presentation of our position more clear, we concentrate
of the method of incremental concept formation proposed by Fisher and
colleges, called
"flexible prediction". In section 2 we recall basic concepts of this method.
In section 3,
recalling our own results, we demonstrate that "relevance"  means 
cheapness in
achieving ones goals. In section 4 we demonstrate the impact of scaling of an
attribute onto its "relevance". In sections 5,6 we show 
the impact of uniform probability distribution
on classification process.
Section 7 shows that pairs of strongly interrelated variables can dominate the
overall classification. Section 8 summarizes our investigation
and presents a proposal
of modification of the approach to clustering tasks. The paper ends with some
concluding remarks.%\\

\section{
Purpose and Method of "Flexible Prediction" Paradigm }

Fisher and colleges \cite{Fisher:85,Fisher:87,%\\
Fisher:87b,Fisher:89,Gennari:89} proposed a new  type of clustering called
"flexible  prediction"., which has been implemented in their COBWEB system
(currently extended in ECOBWEB). 
The goal of clustering via "flexible
prediction"
is to obtain such a hierarchy of concepts  that  the  probability  of 
prediction of sub-class membership from  the  value  of  a  single 
parameter is maximized together with the probability of prediction 
of the value of  the  attribute  from  sub-class  membership.  The 
optimization criterion was: maximize the function \cite{Fisher:87b}:
\begin{equation} 
eval =
 M{^{-1}}\left(
\sum_{C_i}p(C_i) \sum_{A_j\in{\gh D}}
\sum_{V_{jk}} p(A_j=V_{jk}|C_i)^2
-  \sum_{A_j\in{\gh D}} \sum_{V_{jk}} p(A_j=V_{jk})^2
\right)   \label{fisher}
\end{equation}
with:\\
\begin{tabular}[t]{lll}
 {\gh D}&-& set of discrete attributes\\
     $p(C_i)$& -& probability that an object belongs to  
               Sub-category $C_i$\\
     $p(A_j=V_{jk})$ &-& probability that the discrete parameter $A_j$ takes 
               the value $V_{jk}$. \\
    $p(A_j=V_{jk}|C_i)$ &-& respective conditional  probability  within 
               the sub-category $C_i$ \\
      M &-& number of sub-categories.\\
\end{tabular}

The concept hierarchy is developed incrementally, with each new object in a
sequence being classified on the top level into one of the
existing classes or a separate one is created for it depending on which
operation will maximize the above expression. The same is repeated on
sublevels of the hierarchy.%\\

Applicability for diagnosis of plant diseases was claimed \cite{Fisher:89}.%\\

     Gennari et al. \cite{Gennari:89}
extended this methodology to continuous attributes  claiming  success
both for artificial and natural domain  examples  ("forming  diagnostically 
useful categories"). 
The optimization criterion for concept hierarchy was  \cite{Gennari:89}:
\begin{equation}
eval =
 M{^{-1}}\left(
\sum_{C_i}p(C_i) \sum_{A_j\in{\gh K}} 1/\sigma_{A_jC_i}
-  \sum_{A_j\in{\gh K}} 1/\sigma_{A_j}
\right)   \label{gennari}
\end{equation}
with: 
\begin{tabular}[t]{lll}
     {\gh K} &-& set of continuous attributes\\
      $\sigma_{A_j}$  &-& standard deviation continuous parameter $A_j$\\
      $\sigma_{A_jC_i}$ &-& respective standard deviation within the 
               sub-category $C_i$\\
      M &-& number of sub-categories.\\
\end{tabular}

Gennari et al. \cite{Gennari:89} 
 derived their formula - up to a constant factor depending on probability
distribution considered - from Fisher's one
eqn(\ref{fisher}) by substituting unconditional and conditional  probabilities
may be substituted by the respective probability density  functions
and summing with integration.

\section{All Relevant versus Some Relevant}

Developers of "flexible prediction" paradigm noticed that irrelevant
attributes negatively influence the quality of the obtained hierarchy.
One shall however ask what does it mean that the attribute is irrelevant.%\\

In a study, reported in  \cite{Klopotek:91} we assumed that a set of binary
attributes $A_1$,...,$A_n$ (n less than 10) be relevant for a diagnostic task
(previously studied using statistical methods). We made the naive assumption
that if two attributes $A_i,A_j$ are relevant, then their logical conjunction
$A_i\land A_j$, disjuction $A_i\lor A_j$ and negation $\lnot A_j$ should also
be relevant. In this way we built all the anywhere distinct boolean functions
of the relevant attributes. Starting with the set of "relevant" attributes
we tried to construct the classification hierarchy of objects.
 It turned out that the optimization criterion eqn.(\ref{fisher}) drove
each object into a separate case unless two objects were characterized by
exactly the same values of given attributes: any two objects proved to be
equally
similar to one another. The proof that this tendency is an intrinsic property
of the flexible prediction algorithm is given in  \cite{Klopotek:91}. %\\

Clearly, our set of derived attributes contained such absurd functions as
always true and always false ones, but even if one drops them, the situation
does not improve in any way. The lesson from that study is that in order to
obtain a concept hierarchy we need first to ponder some attributes  from the
space of primary and derived relevant attributes. One may do it on
the grounds
of e.g. importance for our target task (more weight for more important
ones) and/or on costs of measurement of a given attribute (more weight for
"cheaper" attributes). But doing so we pre-specify the concept hierarchy
we will derive from our data. And if we would like to get a different concept
hierarchy, a different pondering of the same set of attributes for the very
same set of data will do. We need only to outline our concept hierarchy in
terms of class membership of clustered objects, and then appropiately ponder
the (derived) attributes describing each class (assigning significantly larger
weights to higher hierarchy levels). %\\
 Hence we gain real knowledge from the data only in
that case where we do attribute pondering prior to seeing the data.\\
By the way, this is exactly what statisticians always insist on: define
statistical tests you want to perform before you start collecting the data.%\\

\section{
%2. 
Attributes Equally Pondered and the Effect of Rescaling}
%Implicit Scaling of Attributes}

\Bem{
     We can easily see an  implicit  ponderation  of 
some attributes in eqn(\ref{gennari}). Let us imagine, we  stretch  the  scale
for
the attribute $A_j$ tenfold obtaining $A_j'$  (changing measurement units from
cm   to   mm   e.g.),   then   clearly   $\sigma_{A_j'}=
10*\sigma_{A_j}$
and   hence 
$1/\sigma_{A_j'}=0.1*(1/\sigma_{A_j})$ - so  the  weight  of  an 
attribute  stems from 
stretching   scale.%\\
}

Usually we have no other choice but to decide which attributes are the basis
of the clustering to be performed now. Within this set of attributes it is
natural to assume the same pondering (weighting). We will consider both
optimization criterion (1) and (2) to evaluate them from this point of
view.%\\

It seems that in formula (2) authors pretended to obtain
only a formal and  visual similarity
with (1). To obtain scale-free property, however, the shape of the formula for
continuous attributes should be a little bit different.

\begin{equation} \label{star1}
eval'= \overline{K}^{-1} \sum_{C_i}p(C_i)\sum_{A_j\in K} 
\frac{\sigma_{A_j}}{\sigma_{A_jC_i}}-1
\end{equation}%\\

This formula assures the same result for any linear transformation (i.e.scale)
of attributes which were taken into account.%\\

However, the need of "scale-freedom" effect is not bound to continuous
attributes only. 

     The scaling effect for discrete attributes will  be  demonstrated  by
the following example: 
Let  us  consider  the  attribute  A  taking  6  different  values 
$v_1,v_2,v_3,v_4,v_5,v_6$.  Then  its  contribution  to 
the  evaluation function (eqn.(\ref{fisher})) is:
\begin{equation}
eval_A = \sum_{C_i}p(C_i)\sum_{k=1}^6p(A=v_k|C_i)^2- 
\sum_{k=1}^6p(A=v_k)^2\end{equation}
Let us "rescale" this attribute as to obtain an attribute B  taking 
values $w_1,w_2,w_3$ as follows: B$=w_1$ iff A$=v_1$ or A$=v_6$,  
B$=w_2$ iff A$=v_2$ or A$=v_5$, B$=w_3$ iff A$=v_3$ or A$=v_4$.
Then B contributes:
\begin{equation}
eval_B=\sum_{C_i}p(C_i)\sum_{k=1}^3p(B=w_k|C_i)^2- 
\sum_{k=1}^3p(B=w_k)^2 \end{equation} 
$$=\sum_{C_i}p(C_i)\sum_{k=1}^3p(A=v_k or A=v_{7-k}|C_i)^2- 
\sum_{k=1}^3p(A=v_k or A=v_{7-k})^2$$

Squaring out gives us (taking into account that probability of disjoint
events equals sum of probabilities of the events):
\begin{eqnarray}
\label{cov}
&eval_B =
eval_A+
\sum_{C_i}p(C_i)\sum_{k=1}^32p(A=v_k|C_i)p(A=v_{7-k}|C_i)
\\
&- \sum_{k=1}^32p(A=v_k)p(A=v_{7-k}) \nonumber
\end{eqnarray}

\Bem{
Let us view $p(A=v_k|C_i)$ as a function $f_k(i)$. Then we have
$p(A=v_k)=\sum_{C_i} p(C_i)p(A=v_k|C_i)=E(f_k(I))$. Obviously
$p(A=v_{7-k})=E(f_{7-k}(I))$
so we have:
$$eval_B =
eval_A+  2*\sum_{k=1}^{3} cov_C (f_k(I),f_{7-k}(I))$$

($cov_C$ - covariance of two "attributes" calculated  as  if  the 
set of categories were a sample). 
}

Since $p(A=v_k|C_i)$ for each k=1,2,...,6 can be treated as a function having
as its domain the set of subcategories $\{C_i\}$, one can consider the
following random variable: 
$$Y_k=\Bem{p(A=v_k|.)}=p(A=v_k|C_I)=y_k(I)$$%\\

\noindent
\Bem{
Point in the second expression represents the subcategories with their
respective probabilities while }
I is the indicator random variable. One can
check that expectation of $Y_k$\\
$$E(Y_k)=p(A=v_k), \quad \quad k=1,2,...,6$$%\\
It follows from (\ref{cov}) therefore that % \\
\begin{equation} \label{star2}
eval_A=eval_b-2\sum_{k=1}^3cov(y_k(I),y_{7-k}(I)) %\\
\end{equation}
Let us consider now the reverse rescaling procedure: we start with the
attribute B and afterwards we are trying to split its 3 values into 6. The
intuition suggests us that if way of splitting does not take into account any
information about membership of the record to certain subcategory $C_i$ then
no change should result in the evaluation function.%\\

We model "non-informative splitting" in the following way.%\\

First we randomly generate the record according to probabilities of
subcategories (i.e. $\{p(C_i)\}$). Realization of the random variable I is
given, therefore at this stage. From other point of view a subcategory is
chosen first.%\\

One can observe that evaluation function for discrete attributes does not
depend on any one-to-one transformation of values of attributes. It follows
therefore that "non-informative splitting" can have only the numeric effect if
the probability $$p(B=w_k|C_i)$$ is split in a random way in two parts. %\\

If the splitting is performed in the second stage of random generation
procedure we always would have a negative correlation between the 
components
$$p(A=v_k|C_i)+p(A=v_{7-k}|C_i)$$
of $p(B=w_k|C_i)$.%\\

It follows then from (\ref{star2}) that evaluation function for the attribute
with
more values is higher than for that with 
fewer values. All 3 co variances are
negative. %\\

Such result is difficult to take as rational. The attribute A add no
classification information when comparing with B.%\\

\section{
%3. 
Unbounded  Splitting  of  A  Class  For  Uniformly  Distributed 
Continuous Attributes}

In this section we will consider somewhat ideal clustering situation in the
case of continuous attributes.%\\

Assume that each attribute $A_j$ takes values in the interval of the length
$\Delta_j$. The behaviour of attributes is very positive and clear from the
point of view of distinguishing subcategories $C_i$. This behaviour is 
formalized in the following way.%\\

Each attribute for the whole population has the uniform distribution within
its interval. For the subcategories attributes take values only within
intervals of the lengths which corresponds to probability of the
subcategory. It means that attribute $A_j$ for subcategory $C_i$ has values
within the interval of length: $$\Delta_j\cdot P(C_i)$$ %\\
For  given  attribute  open   subintervals   
which represent  M 
subcategories are
assumed to be disjoint. The value of attribute gives therefore the sure
classification of the object (record) which corresponds to this value. %\\

To conclude the description of the above ideal situation one  can 
notice
that distribution for subcategory (i.e., conditional distribution for records
within $C_i$) must be uniform either.%\\

The elementary calculations lead to the following simple expression of
evaluation function (2) and (\ref{star1}):%\\

\begin{equation}  \label{star4}
eval_{in(2)}=\frac{(M-1)2\sqrt{3}}{M}\sum_{A_j\in K}\frac{1}{\Delta_j}
\end{equation} 
\begin{equation}  \label{star3}
eval_{in(\ref{star1})} = M-1
\end{equation} 
The second result for the idealized clustering situation seems to be more
rational than for the first one. The proposed evaluation function 
depends neither on the domains of attributes nor on their number. Actually
the entire clustering information is included in one attribute. All
attributes are copies of each other from this point of view.%\\

Of course the lengths $\Delta_j$ have nothing to do with precision of
clustering. The precision is higher, however, for the attribute which
classifies 3 subcategories than for attribute classifying 2 clusters. So it
seems rational that the form (\ref{star3}) has stronger relationship with M
than (\ref{star4}). %%\\

\section{
%3. 
Shift of Mass Between Classes of Different Density}

Let us  check  one  more  aspect  of  attribute  behavior.  Let  us 
approximate the  overall  attribute  distribution  with  pieces  of 
uniform  density  (e.g.  a  histogram).   Let   us   examine   the 
optimization tendency of the splitting criterion.\\
Let us consider the following experiment:\\
Let us take two "neighboring" classes $C_1,C_2$:  
two uniformly distributed neighboring intervals (densities $D_1$  and 
$D_2$ respectively) with lengths b and a.\\
The contribution of these intervals to the evaluation is:\\
      $p(C_1)=D_1*b, p(C_2)=D_2*a, 
\sigma_{AC_1}=(2\sqrt{3})^{-1}b,
\sigma_{AC_2}=(2\sqrt{3})^{-1}a$, hence: % (eqn.(\ref{gennari})):
$$
M{^{-1}}\left(\left(p(C_1)/\sigma_{AC_1}+p(C_2)/\sigma_{AC_2}\right)
-1/\sigma_A \right)=
$$
\begin{equation}
M{^{-1}}\left(\left(D_1*b/((2\sqrt{3})^{-1}b)   
+D_2*a/((2\sqrt{3})^{-1}a)\right)
-1/\sigma_A \right)=
\end{equation}
$$
M{^{-1}}\left(\left(D_1*2\sqrt{3}   
+D_2*2\sqrt{3}\right)
-1/\sigma_A \right)
$$ %\end{equation}

This actually means the following: the contribution of a attribute depends 
on the densities of intervals and not on their widths. 
 We want next to optimize the criterion by shifting the
boundary between the intervals increasing the class $C_1$ at  the 
expense of
the class $C_2$ so that $C_2$ consists only of an interval with width
a-x and density $D_2$, and $C_1$ consists of an interval with length $b$ and
density $D_1$ and an interval $x$ with density $D_2$. 
It is obvious that the contribution of $C_2$ to the evaluation function will
remain the same so we need only to consider the class $C_1$. We will try to
find out what value shall be taken by $x$.%\\

For consideration of standard deviation within class $C_1$ we need to rescale
the densities $D_1,D_2$ of subintervals b and x to $d_12,d_2$ so that
$d_1b+d_2x=1$. 
Variance of $C_1$ then equals to: 
$$ %
\sigma_{AC_1}^2=
\int_{-b}^{0}d_1y{^2}dy+\int_{0}^{x}d_2y{^2}dy-
\left( \int_{-b}^{0}d_1ydy+\int_{0}^{x}d_2ydy-  \right)^2=$$
$$=         d_1b{^3}/3+d_2x{^3}/3-(-d_1b{^2}/2+d_2x{^2}/2)^2=
=d_1 \frac{b{^3}}{3} +d_2 \frac{x{^3}}{3} 
-d_1^2 \frac{b{^4}}{4} -d_2^2 \frac{x{^4}}{4} +
2d_1d_2 \frac{b{^2}x{^2}}{4} $$

We introduce the coefficient $q$ such that  $d_2=qd_1$, then
$d_1(b+qx)=1$, Just $d_1=\frac{1}{b+qx}$, $d_2=\frac{q}{b+qx}$. 
With this substitution we get:
$$
\sigma_{AC_1}^2=
=\frac{1}{b+qx} \frac{b{^3}}{3} +\frac{q}{b+qx} \frac{x{^3}}{3} 
-\frac{1}{(b+qx)^2} \frac{b{^4}}{4} -\frac{q{^2}}{(b+qx)^2} \frac{x{^4}}{4} +
2\frac{q}{(b+qx)^2} \frac{b{^2}x{^2}}{4} $$

Probability of class $C_1$   is now $p_{C_1} = D_1*b+D_2*x$ with
$D_2/D_1=d_2/d_1=q$, hence \begin{equation}
p_{C_1} /\sigma_{AC_1} =\frac{D_1}
{\sqrt{
\frac{1}{(b+qx)^3} \frac{b{^3}}{3} 
+\frac{q}{(b+qx)^3} \frac{x{^3}}{3} 
-\frac{1}{(b+qx)^2} \frac{b{^4}}{4} 
-\frac{q{^2}}{(b+qx)^4} \frac{x{^4}}{4} +
2\frac{q}{(b+qx)^4} \frac{b{^2}x{^2}}{4} 
}}\end{equation}

To maximize the quotient,  the  denominator,  hence  the  squared 
denominator must be minimized. The derivative
$d(squared denominator)/dx  =  q*(1-q)*(b+qx)^{-5}b{^2}x(b+x)$. 
   will   be 
negative (hence steadily falling with increase of x) iff q exceeds 1 which
means $d_1
<d_2$ , which means that lower density intervals will 
swallow 
 higher density
ones. %\\
We see that mutual prediction tends to isolate extremely narrow
modes from a multimodal distribution. We have here again a lesson that
discretization of the predicted attribute is essential.%\\

\section{
%4.
 Overfitting A Continuous Attribute}
Let us consider again the discrete mutual prediction function of 
\cite{Fisher:89} for a selected attribute A and the classifying variable C:
$$\sum_x\sum_yp(A=y)*p(A=y|C=x)*p(C=x|A=y)=
\sum_x\sum_y\frac{p(A=y \land C=x)^2}{p(C=x)}$$

Let  us  generalize  it  to   a   continuous   "classification" 
variable C and continuous attribute A connected by a bivariate 
normal distribution:

\begin{eqnarray}
&\int_{-\infty}^{\infty}\int_{-\infty}^{\infty}
\frac{
\left(
\frac{1}{2\Pi\sqrt{1-{R}^2}\sigma_x\sigma_y}
\exp\left(
\frac{1}{2(1-{R}^2)}
      \left(
      (\frac{x-\mu_x}{\sigma_x})^2
-2R
      (\frac{x-\mu_{A_j}}{\sigma_x})
      (\frac{y-\mu_y}{\sigma_y})
+
      (\frac{y-\mu_y}{\sigma_y})^2
      \right)
     \right)
\right)^2
}
{
\frac{1}{\sqrt{2\Pi}\sigma_x}
\exp\left(\frac{1}{2} (\frac{x-\mu_x}{\sigma_x})^2\right)
}
dxdy= \nonumber\\
&= \left(   2\sigma_y\sqrt{\Pi}\sqrt{1-2R{^2}}\right)^{-1}
\end{eqnarray}

The resulting formula points at  the following:
\begin{itemize}
\item  Dependence on attribute scale $\sigma$ -  the  larger  the  range  of 
attribute values, the smaller attribute variance  the  higher  its 
importance
\item Dependence   on   correlation:   the   stronger   the 
correlation  (closer  to  0.7!!!!)  the   higher   the   attribute 
importance.
\end{itemize}
REMARK: for correlations over 0.7 - nonsense values. 0.7*0.7=0.49.
- if C gets  closer  to  A  (higher  correlation)  then  the  term 
concerning A in 
Fisher's evaluation function will absolutely  dominate  the  whole 
expression, hence overfitting towards  one  of  the  attributes is 
guaranteed. %\\

We shall conclude from this case study in "continuous classification" that
strongly interrelated attributes can dominate the classification hierarchy so
that they will disable it to predict values of other attributes of interest.
Therefore when designing an evaluation function for a set of relevant
attributes, if we want to have a balanced mutual prediction capability, 
 correlation between the classification and each attribute of
interest should negatively influence the weight of that attribute in the
classification evaluation function.%\\

\section{Discussion}

In this paper the flexible prediction (FP) paradigm for knowledge discovery
from a set of cases has been studied. It has been demonstrated that:%\\
\begin{itemize}
\item 
 the optimization criterion of FP drives 
each object into a separate case unless two objects were characterized by
exactly the same values of given attributes: any two objects proved to be
equally
similar to one another, if we take  primary relevant attributes and
all their derived attributes, %\\
\Bem{
\item
 a more precise 
attribute value
measurement procedure (with a more refined set of possible attribute values)
may 
cause it to deteriorate the optimization function and hence to contribute
less to the overall classification. %\\
}
\item
for uniformly distributed continuous attributes the optimization 
 function behaves in a not optimal way
\Bem{
  causes unlimited growth of the number of  
sub-classes in a classification hierarchy while 
the splitting point  positions between classes do  not  influence  the 
evaluation function, especially no tendency  towards  even  length 
intervals can be observed}%\\
\item
mutual prediction paradigm of COBWEB evaluation function
tends to isolate extremely narrow
modes from a multimodal distribution for continuous attributes%%\\
\item 
the importance of an attribute in the optimization depends on its scale
in an unwanted way
\Bem{:
 the  larger  the  range  of 
attribute values, the less important the attribute may be for discrete
attribute, if the increased range is an effect of random splitting of
attribute values%\\
\item 
the importance of an attribute in the optimization depends on its variance:
the smaller attribute variance  the  higher  its 
importance }
\item 
the importance of an attribute in the optimization depends on its correlation
with the classification criterion:
  the   stronger   the 
correlation  the   higher   the   attribute 
importance.
\end{itemize}

%\\

Therefore :

\begin{itemize}
\item 
in order to
obtain a useful
 concept hierarchy we need first to ponder those attributes  from the
space of primary and derived relevant attributes, which are easily measurable
and/or interesting ones%\\
\item 
if we want to have mutual prediction:
classification $<->$ attribute value, we should treat both directions of
"implication" asymmetrically: On the one hand, implication  classification
$<-$ attribute value should exploit as high precision of attribute measurement
as available, on the other hand the implication classification $->$ attribute
value should accept attribute value with as low precision as satisfactory.%\\
\Bem{
\item
 when designing an evaluation function for a set of relevant
attributes, if we want to have a balanced mutual prediction capability, 
 correlation between the classification and each attribute of
interest should negatively influence the weight of that attribute in the
classification evaluation function.%\\
}%\\
\end{itemize}

%\\

\subsection{ A Possible Recovery}

To meet the above requirements a redesign of the evaluation function for
flexible prediction seems to be necessary. 
We shall handle below only the case of discrete attributes.
First of all one shall not insists
that all the attributes need to be predicted from the evaluation function. 
Hence 
 the evaluation
function should consist of two parts: one concerning the predicted attributes
and one concerning predicted attributes.%\\

As the contribution of predicting attributes to the evaluation function of the
classification hierarchy is concerned, one shall construct a
function $f_A:A\rightarrow C$ ($A$ - the attribute, $C$ the classification) in
such a
way as to ensure that $p(C=f_A(A))$ (that is the probability that the
attribute
A predicts the classification C correctly) is maximized. Then the measure of
contribution of the attribute A to the evaluation of  classification
would be  $p(C=f_A(A))$. Please pay attention to the fact that if we refine
the
attribute A (increase the precision of measurements), but this increase does
not improve the quality of prediction of the class, then the value of
$p(C=f(A))$ will not change (it will not decrease). The only case when
$p(C=f(A))$  may decrease is when the number of classes in classification C is
increased. In this way the symmetry, anchored in the correlation measures,
between the predicting attribute and the classification variable is broken.%\\

As the contribution of predicted  attributes to the evaluation function of the
classification hierarchy is concerned, one shall construct a
function $g_A:C \rightarrow A$ (A - the attribute, C the classification) in
such a way as to ensure that $p(A=g_A(C))$ (that is the probability that the
attribute A is predicted by classification C correctly) is maximized. Then
the measure
of contribution of the attribute A to the evaluation of  classification
would be  $p(A=g_A(C))$. Please pay attention to the fact that if we refine
the classification $C$
 (increase the number of classes), but this increase does
not improve the quality of prediction of the class, then the value of
$p(A=g_A(C))$ will not change (it will not decrease). The only case when
$p(A=g_A(C))$  may decrease is when the required precision of prediction of A
is required. In this way the symmetry, anchored in the correlation measures,
between the predicted  attribute and the classification variable is broken.%\\

Last not least the case of an attribute A shall be considered which shall both
be predicted from the classification and if available predict the
classification. We must then construct two attributes: A' and A" out of it
constructing two functions: $f_{A'}:A'\rightarrow C$ and  
$g_{A"}: C \rightarrow A"$
according to
the rules indicated above. The attribute A" should in general have a different
precision (grid of values) that A': A' should be finer than A" so that A' is
measured as precisely as possible and A" should be predicted only as precisely
as necessary. %\\

In the end, we shall consider the overall structure of the evaluation
function. Please recall the fact that the flexible prediction paradigm is
used in combination with incremental concept formation that is with 
each new incoming case the classification hierarchy is re-evaluated.
We propose the following evaluation function for each step of the process:%\\
\begin{equation}
\sum_{A\in Prtd} w_A  \frac{p(A=g_A(C))}{p'(A'=g'_A(C))}
\end{equation}
$$
+
w_C \sum_{A\in Prting}
\frac{ \max_A(p(A available)\cdot p(C=f_A(A)))}
{ \max_A(p(A available)\cdot p'(C'=f'_A(A)))}
$$
where p', C', f' and g' are probabilities, classification and prediction
functions used on occasion of the previous case incorporated into the
hierarchy, p(A available) is the probability that an attribute is available
for  observation, coefficients w are weights assigned to knowledge of
predicted attributes and the classification, Prtd is the set of predicted
attributes, Prting is the set of predicting attributes. The motivation behind
the proposal is the following:\\
1. We maximize over probability of correct prediction of classification
(weighed by probability of availability) in order to eliminate impact of
irrelevant attributes: irrelevant attributes will simply not predict the
classification strongly enough.\\
2. We relate predictability of each predicted attribute and of the
classification to predictability of the previous step via quotients in order
to balance the predictability among those attributes so that the
classification does not tend to favor any of the predicted attributes. In
this way a too strong correlation between classification variable and any of
the predicted attributes at expense of low correlation with other
predicted attributes is prevented\\
3. By regulating proportions between weights of classification variable and
predicted attributes we can refrain the classification from splitting into too
much classes. This is because the $p(C=f_A(A)$ gets lower if we split C beyond
predictability from predicting attributes. On the other hand predicted
attributes will always profit from split of classification  because
$p(A=g_A(C)$ never gets lower with refinement of C, so refinements of C are
favored even if only prediction of few of predicted attributes can be
improved from the refinement of C. %\\

Please notice that attachment of transformations of the space of predicting
attributes (e.g. combining two attributes to one) will not deteriorate the
performance of the evaluation criterion, we need only to calculate the
probability of availability of the newly generated attribute.%\\

The study carried out so far on this formula has been theoretic in nature and
empirical tests on artificial data indicate necessity of development of
stabilizing strategies for the starting stages of the concept formation
process (e.g. replacement of maximum by "take n largest values"
function).
Furthermore, extensions to continuous case imply important questions
of discretization strategies both for predicted and predicting attributes.
However, we feel that the departure from symmetry of predicting and predicted
attributes is a promising path of research, including split of attributes
which are both predicted and predicting into two separately treated
attributes. Also balancing of prediction of
attributes from the classification variable seems to be important improvement
of the flexible prediction paradigm. %\\

We feel that usage of a kind of cost function instead of probability of
availability may also be profitable.%\\

\section{%7.
 Concluding Remarks}

The flexible prediction paradigm for search of knowledge about relevant
classification hierarchies has the following drawbacks:%\\
\begin{itemize}
\item is negatively influenced by irrelevant attributes and 
may be badly influenced by abundance of relevant attributes,
\item may be negatively influenced by increase of attribute precision, 
\item must be optimized for piecewise uniform continuous
attributes, \item is sensitive to attribute scaling, variance and correlation
\end{itemize}
To achieve a balanced and useful classification, following 
modifications of the
paradigm are necessary:
\begin{itemize}
\item attributes must be split into three categories: those which should
be predicted from the hierarchy, those which are predicting the classification
hierarchy and those having both roles%\\
\item all three categories need separate treatment: %\\
\begin{itemize}
\item predicted attributes should be carefully selected based on their 
utility,
 the domain of predicted attribute values should be as restricted as
possible (precision as low as possible)%\\
\item attributes only predicting the classification have no such restrictions
- they may be as many as available and as precise as available,
however they should be pondered depending on expenses of their
measurement and possibility of their measurement.%\\
\Bem{
\item attributes both predicting and predicted should be treated as two
distinct attributes: one predicting and one predicted, each of them
obeying the rules given above%\\
}
\end{itemize}  
\item correlation between the classification and a predicted attribute 
 should negatively influence the weight of that attribute in the
classification evaluation function.%\\
\end{itemize}
%\\

\end{document}